\documentclass[10pt,twocolumn,letterpaper]{article}

\usepackage{cvpr}              
\usepackage{multirow}
\usepackage{booktabs}
\usepackage{enumitem}

\definecolor{cvprblue}{rgb}{0.21,0.49,0.74}
\usepackage[pagebackref,breaklinks,colorlinks,allcolors=cvprblue]{hyperref}

\newcommand{\pt}{\phantom{0}}
\newcommand{\ptt}{\phantom{00}}

\def\paperID{10184} 
\def\confName{CVPR}
\def\confYear{2025}

\title{Diffusion Product Quantization}

\author{Jie Shao \quad\quad Hanxiao Zhang \quad\quad Jianxin Wu\\
National Key Laboratory for Novel Software Technology, Nanjing University, China\\
School of Artificial Intelligence, Nanjing University, China\\
}

\begin{document}
\maketitle

\begin{abstract}
	In this work, we explore the quantization of diffusion models in extreme compression regimes to reduce model size while maintaining performance. We begin by investigating classical vector quantization but find that diffusion models are particularly susceptible to quantization error, with the codebook size limiting generation quality. To address this, we introduce product quantization, which offers improved reconstruction precision and larger capacity—crucial for preserving the generative capabilities of diffusion models. Furthermore, we propose a method to compress the codebook by evaluating the importance of each vector and removing redundancy, ensuring the model size remaining within the desired range. We also introduce an end-to-end calibration approach that adjusts assignments during the forward pass and optimizes the codebook using the DDPM loss. By compressing the model to as low as 1 bit (resulting in over 24× reduction in model size), we achieve a balance between compression and quality. We apply our compression method to the DiT model on ImageNet and consistently outperform other quantization approaches, demonstrating competitive generative performance.
\end{abstract}

\section{Introduction}
\label{sec:intro}

\begin{figure}
   \centering
   \includegraphics[width=0.85\linewidth]{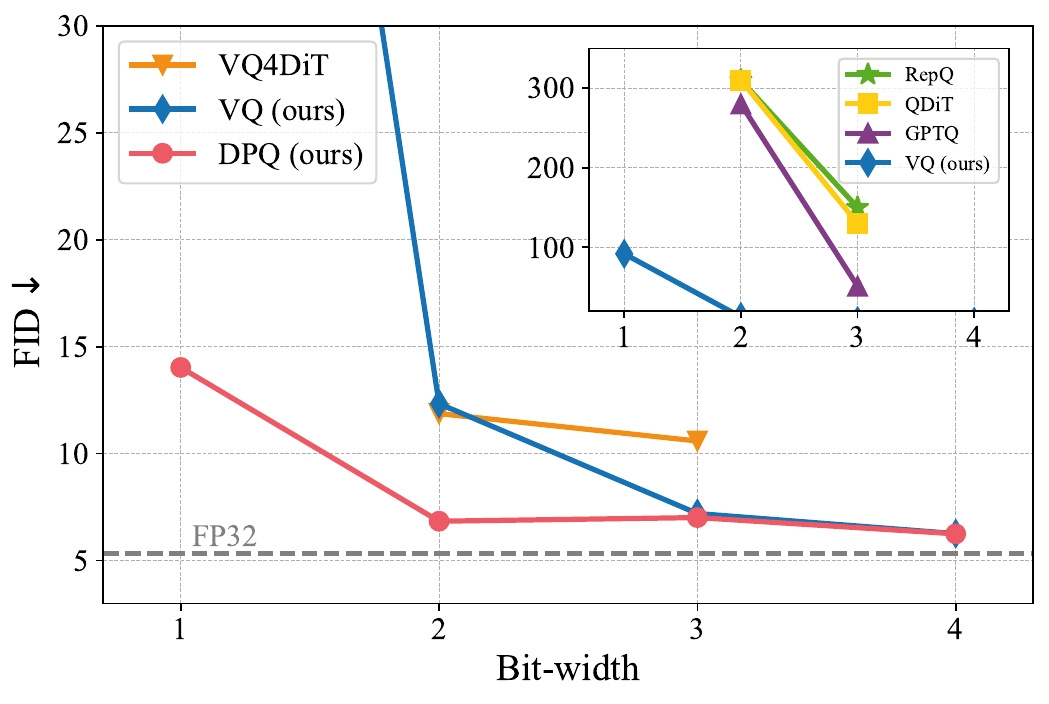}
   \caption{Compression results at low bit-widths, using the DiT-XL/2 model with 250 sampling steps and CFG of 1.5.} 
   \label{bitwidth}
\end{figure}

Deep learning has achieved remarkable success across a wide range of vision and language tasks~\cite{resnet, clip, gpt3}. Recent progresses in diffusion models have enabled the generation of high-quality images, positioning them as a leading approach in image-level generative modeling~\cite{ddpm, score-sde, guided-diffusion}. Despite their impressive performance, the large number of parameters and high computational complexity of diffusion models make them too resource-intensive for practical deployment. This presents significant challenges for their use in \textit{resource-constrained} environments.

To address the challenges of deploying diffusion models, compression techniques, particularly quantization~\cite{ptq4dm, ptqd, qdit}, have become crucial for accelerating and reducing the size of these models. Quantization reduces the precision of weights and activations by converting them from floating-point values to lower-precision formats, thereby preserving the generative quality of diffusion models while significantly lowering resource consumption and inference time. Among these techniques, vector quantization (VQ)~\cite{vq} has proven to be an effective method for compressing weights to \textit{extremely low bit-widths}. VQ methods partition model weights into sub-vectors, cluster them and form a codebook using the clustered centroids, then any sub-vector is \emph{only} stored by the index of its nearest neighbor in the codebook. This approach allows for significant compression of a model, resulting in substantial reductions in model size across various architectures~\cite{AndTheBit, gptvq, vq4dit}. We implemented VQ compression for DiT, with results aligning closely with~\cite{vq4dit}. VQ achieves a \textit{$16\times$} reduction in model size while preserving the ability to generate meaningful images, but uniform quantization fails to perform effectively at such low bit-widths, as illustrated in~\cref{bitwidth} in the smaller window at the top-right corner.

But, VQ faces limitations in high-dimensional contexts~\cite{pq}, as it becomes increasingly difficult to accurately represent high-dimensional vectors with a shared, limited-length codebook. This issue becomes more pronounced as the vector size increases (\ie, when the bit-width decreases), making precise quantization at low bit-widths infeasible and leading to significant reconstruction errors. Our experiments in~\cref{bitwidth} reveal two major issues when applying VQ to the compression of diffusion models. First, as the bit-width decreases, the Fréchet Inception Distance (FID) increases significantly, particularly when reducing from 2 bits to 1 bit. Second, while VQ outperforms uniform quantization, the performance degradation remains substantial. The first issue arises from the inherent difficulty of \textit{quantizing high-dimensional sub-vectors} in VQ. The second issue stems from the nature of diffusion models themselves. Unlike typical models which generate output in one forward pass, the diffusion process involves multiple time steps of forward computations. During this iterative process, \textit{quantization errors accumulate over time}, further degrading the quality of the generated outputs. Thus, to better align with the characteristics of diffusion models, a new quantization method that can effectively handle high-dimensional vectors with higher precision is required.

To overcome both difficulties, we revamp and redesign a classic quantization technique, product quantization (PQ)~\cite{pq, opq}, which is commonly used for approximate distance computation, to compress diffusion models. We refer to our approach as \textit{Diffusion Product Quantization} (DPQ). The fundamental concept of PQ is to split the original vector space into a Cartesian product of several low-dimensional subspaces, with each subspace being quantized into a set of codewords~\cite{opq}. For diffusion models, we decompose the row vectors into multiple low-dimensional sub-vectors, where each column of sub-vectors at the same position shares a \textit{separate} codebook. By utilizing different codebooks for each subspace, the effective number of codewords in the original space grows \textit{exponentially}, while the increase in codebook size remains linear. This approach enables both high-dimensional representation and precise quantization, making it possible to achieve extreme low-bit compression of diffusion models down to 1 bit.

A key difficulty in adapting PQ for diffusion models is that \emph{the overhead outweighs benefits}, more precisely, the reduction in model weights' storage size is quickly canceled by the linear increase in codebook sizes (which have to be stored, too). In DPQ, we propose to \emph{compress the codebook itself}, too. We start by sorting the centroids in each codebook according to their usage frequency, and then compress them by calculating the distances between centroids. This compression prioritizes the use of more frequent centroids while eliminating redundancy.

Finally, we calibrate the codebooks and assignments in an end-to-end manner. During the forward of the diffusion model, we update the PQ assignments based on the quantization error of activations, and during the backward, we fine-tune the codebook using the DDPM loss. As a result, DPQ consistently outperforms VQ by very large margins at all bit-widths, demonstrating superior performance across the entire compression range, as shown in~\cref{bitwidth}.

The contributions are summarized as follows:
\begin{itemize}[leftmargin=2em]
   \item We propose DPQ, an effective framework that pioneers product quantization for extremely low bit-width compression of diffusion models.
   \item We introduce a novel method for compressing the codebooks themselves by ranking centroids based on their usage frequency and selectively compressing centroids based on similarity. We also propose a calibration approach that updates assignments using activations in the forward pass and refines the codebooks in the backward pass. 
\end{itemize}
Our method demonstrates superior performance in compressing DiT, achieving 1-bit compression and a model size reduction of over $24\times$. It outperforms all other compression methods in low-bit-width settings.

\section{Related Work}
\label{sec:relatedwork}

\begin{figure*}
    \centering
    \includegraphics[width=\linewidth]{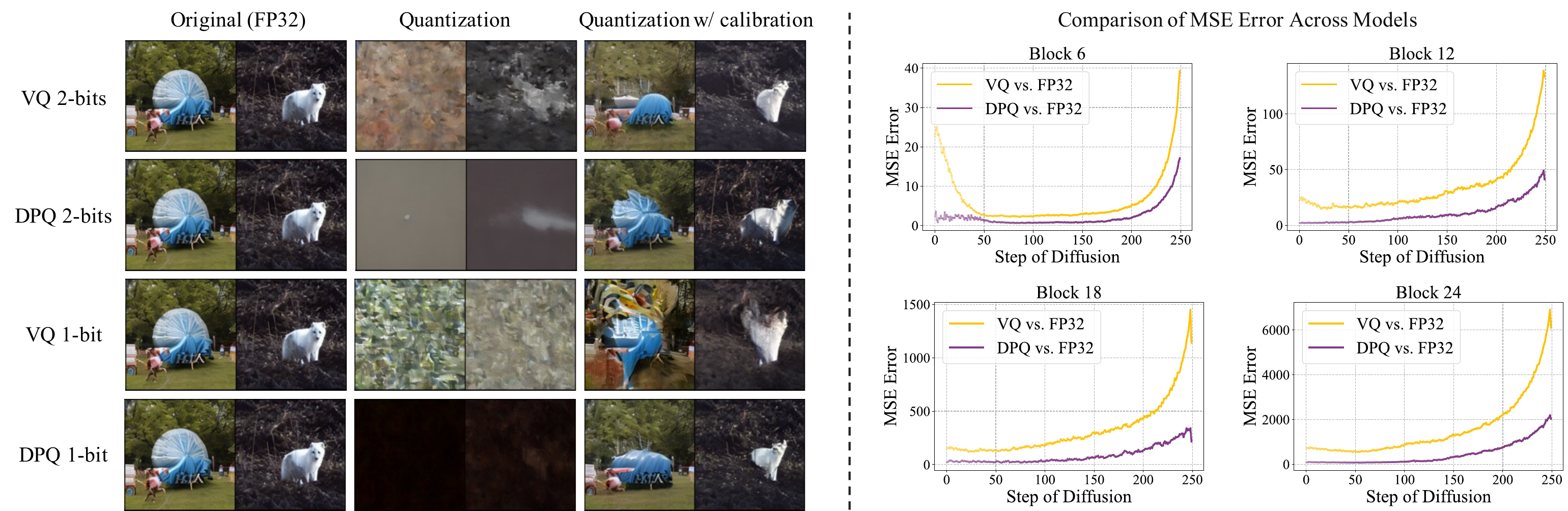}
    \caption{Results of VQ compression at 2 bits and 1 bit. The VQ-compressed model shows degradation and noticeable distortions, which become more severe as the bit rate decreases. In contrast, our DPQ method largely preserves the model's capacity (left). Additionally, VQ causes quantization errors in each block's output to accumulate over diffusion steps, whereas our DPQ method effectively limits quantization error within a controlled range (right).}
    \label{fig:compare}
\end{figure*}

\paragraph{Diffusion Models}
Diffusion~\cite{dpm, ddpm} and score-based generative models~\cite{score-matching, sde}, which have surpassed generative adversarial networks (GANs)~\cite{gan} as the state-of-the-art in image generation, demonstrate exceptional success in producing high-quality images~\cite{glide, dalle2, ldm}. From an architectural standpoint, UNet-based diffusion models have long dominated image generation tasks. However, recent research has increasingly shifted toward transformer-based architectures~\cite{diffusion-survey} and led to the development of Diffusion Transformers (DiTs)~\cite{dit}, which set new benchmarks in image generation performance. The scalability gives DiTs significant potential to expand their applicability across a broad range of generative tasks~\cite{dit, dit3d, opensora}.

Given the substantial computational resources required by diffusion models, deploying them on resource-constrained devices is challenging. To address this, we propose a novel quantization method for extreme low-bit-width compression of diffusion models, which significantly reduces their memory footprint.

\paragraph{Quantization for Model Compression}
Model quantization is widely used to reduce model size and accelerate inference by converting weights and activations from high-precision floating-point values to lower-precision representations. Extensive research has been conducted on uniform quantization for transformer-based models across various tasks and model scales. GPTQ~\cite{gptq} introduces an approximate second-order method that leverages Hessian information, while AWQ~\cite{awq} uses per-channel scaling to minimize the quantization loss for critical weights.

In uniform quantization, each scalar value in weights or activations is quantized individually, typically compressed to 8 bits. However, Vector Quantization (VQ)~\cite{vq} offers a more efficient quantization technique for compressing model weights, enabling more flexible and adaptable representations. Drawing from concepts similar to \textit{k}-means clustering, VQ has been applied to transformers in various contexts~\cite{sparse-attention, product-key, quip, aqlm}. GPTVQ~\cite{gptvq} performs vector quantization of the weight tensor in a greedy manner, while VQ4DiT~\cite{vq4dit} designs a zero-data, block-wise calibration process to progressively refine each layer's codebook. Although the VQ method has been successfully applied to large language models (LLMs), its application to diffusion models is more challenging due to the high dimensionality of the sub-vectors and the accumulation of quantization errors in the diffusion process.

\paragraph{Product Quantization}
Product quantization (PQ) was originally developed for nearest-neighbor search~\cite{pq,opq,ck-means} as an efficient solution to address the challenges of high-dimensional vector quantization. Unlike traditional vector quantization, which becomes infeasible for very large codebooks, PQ reduces complexity by partitioning the vector into smaller sub-vectors, each of which can be quantized separately. This technique allows flexibility in choosing the length of sub-vectors and the number of sub-vector groups to be quantized jointly~\cite{AndTheBit}. In our paper, we revamp PQ for diffusion models, introducing a codebook compression method to reduce the unacceptable memory footprint caused by the codebooks, along with a calibration technique that dynamically updates PQ assignments and codebooks during model training, enabling efficient and accurate quantization with minimal impact on model performance.

\section{Method}
\label{sec:method}

We start by briefly introducing the preliminaries.

\textbf{Vector Quantization.} Let $\mathbf{W} \in \mathbb{R}^{m\times n}$ represents the input to the quantizer, where $m$ and $n$ are the number of input and output channels, respectively. Vector quantization discretize a floating-point vector to an integer index to a centroid within a codebook $\mathbf{C} = \{\mathbf{c}_1, \mathbf{c}_2, \dots, \mathbf{c}_k\}$. Each high-precision vector in $\mathbf{W}$ is represented by an index $i$ pointing to a centroid $\mathbf{c}_i$. This index can be stored with $\lceil \log_2 k \rceil$ bits, allowing for efficient compression of weight tensors by choosing $k$ such that $\log_2 k$ is less than the original bit width of $\mathbf{W}$. The codebook is typically trained using K-Means clustering, and the assignment is calculated by finding the nearest centroids.

A vector quantizer can also be constructed by setting a sub-vector dimension for the centroids in the codebook $\mathbf{C}$. VQ partitions $\mathbf{W}$ into row sub-vectors $\mathbf{w}_{i,j} \in \mathbb{R}^{1 \times d}$, where $1 \leq i \leq m$ and $1 \leq j \leq n/d$, with $d$ being a divisor of $m$. This setup results in a total of $m \times n/d$ sub-vectors. These sub-vectors are quantized using a codebook $\mathbf{C} \subseteq \mathbb{R}^{k \times d}$. The original weight $\mathbf{W}$ can then be expressed by an assignment $\hat{\mathbf{W}}$, whose elements are calculated as
\begin{equation}
    \hat{w}_{i,j} \triangleq\mathop{\arg\min}\limits_{p} \|\mathbf{w}_{i,j} - \mathbf{c}_p\|^2 \,,
\end{equation}
which represents the index of each codeword in the codebook that best fits each sub-vector $\mathbf{w}_{i,j}$. The assignments require $m \times n/d \times \lceil \log_2 k \rceil$ bits for storage, while the codebook itself requires $k \times d \times 32$ bits. The overall compression ratio of $\mathbf{W}$ is $d \times \lceil \log_2 k \rceil$, where we neglect the codebook's storage because it is small. Assuming the original format is FP32, the bits per value after quantization become $\frac{32}{d \lceil \log_2 k \rceil}$ bits.

\textbf{Diffusion Model.} Diffusion model is the most advanced generative model that can generate high-quality images. At every
training step, the model sample a mini-batch of images $x_0$ from the dataset and add noise to obtain $x_t$: $ q(x_t \mid x_0) = \mathcal{N}(\sqrt{\bar{\alpha}_t} x_0, (1 - \bar{\alpha}_t) I)$. After training a diffusion model $\theta$ to get $p_{\theta}(x_{t-1} \mid x_t)$, we reverse the above process step by step to recover the original image $x_0$ from pure noise $x_T \sim \mathcal{N}(0, I)$.
The training objective is to reduce the gap between the added noise in the forward process and the estimated noise in the reverse process:
\begin{equation}
    \label{eq:ddpm}
    L_{\text{DDPM}} = \mathbb{E}_{t \sim [1, T]} \| \epsilon_t - \epsilon_{\theta} (\sqrt{\bar{\alpha}_t} x_0 + \sqrt{1 - \bar{\alpha}_t} \epsilon_t, t) \|^2,
\end{equation}
where $\epsilon_t \sim \mathcal{N}(0, I)$ is the noise added to original images and $\epsilon_{\theta}$ is the noise estimated by the
trainable model.

\begin{table}[tb]
    \centering
    \resizebox{\linewidth}{!}{%
    \begin{tabular}{l|c|c|c|c|c}
        \toprule
        Method & bit-width & FID$\downarrow$ & Size ratio & $\mathbf{C}$ ratio & $\hat{\mathbf{W}}$ ratio \\
        \midrule
        VQ4DiT & 2 & 11.87 & 15.75$\times$ & - & - \\
        \midrule
        VQ (1 ep) & 2 & 12.33 & 14.87$\times$ & 4666$\times$ & 16$\times$ \\
        VQ (10 ep) & 2 & 12.12 & 14.87$\times$ & 4666$\times$ & 16$\times$ \\
        DPQ & 2 & \pt6.84 & 12.08$\times$ & \ptt71$\times$ & 16$\times$ \\
        \hline
    \end{tabular}%
    }
    \caption{Comparison of compression methods with their FID and compression ratios. Size ratio, $\mathbf{C}$ ratio and $\hat{\mathbf{W}}$ ratio indicate the ratio of the original model size to the compressed model size, codebook size and compressed weights size, respectively. Note that the compressed model size is the sum of codebook size and compressed weights size.}
    \label{tab:vq}
\end{table}

\subsection{VQ compression for Diffusion}

To validate the effectiveness of VQ compression, we first apply a simple VQ method to the DiT model. We quantize the weights of all linear layers in each DiT block, excluding the bias terms, as they are much smaller than the weights. Following a common VQ approach, we quantize the weights in groups. Specifically, we partition the weights into groups with each sub-vector dimension $d=4$. We use INT8 quantization, meaning each sub-vector $\mathbf{w}_{i,j}$ in each group is quantized to an INT8 integer. Consequently, the weights $\mathbf{W}$ are quantized to $\hat{\mathbf{W}}$, stored as a codebook with a shape of $256 \times 4$ and an assignment matrix $\mathbb{N}^{m \times \frac{n}{4}}_{0}$, computed using K-means with 1000 iterations. This results in a $16\times$ compression of the model, or equivalently, 2-bit compression from the 32-bit FP32.

Directly applying VQ compression to the DiT-XL/2 model results in a significant performance drop. As shown in~\cref{fig:compare}, this compression causes the model to generate images that are meaningless and resemble pure noise. A solution is to calibrate the compressed model. Since the assignments are fixed, we can directly apply~\cref{eq:ddpm} to recalibrate the model. After training for one epoch, the generative performance improves significantly. We sample 10,000 images from the compressed model with 250 sampling steps and a CFG of 1.5, and calculated the FID, with results listed in~\cref{tab:vq}. We also compare with recent work that uses VQ for DiT compression~\cite{vq4dit}---our results are comparable despite some differences in implementation details. This indicates that VQ successfully compresses the DiT model by 16 times while retaining its image generation capabilities.

But, compared to VQ-compressed large language models (LLMs)~\cite{quip, aqlm, gptvq}, the performance degradation in DiT appears much more significant. We also experiment with 1-bit compression, but the diffusion model's performance deteriorated substantially (the compression size ratio is not exactly $32\times$ due to the increased relative proportion of uncompressed components, such as the final layer). This highlights the need to understand the bottlenecks in compressing diffusion models.

To investigate further, we visualize the L2 distance between the features of the VQ-compressed model and the original model. We arbitrarily select block 6, 12, 18 and 24, positioned midway through the model, to examine its features before and after quantization. The average reconstruction loss of these features is shown in~\cref{fig:compare} (right part). Notably, even after calibration, features of the quantized block deviate significantly from those of the original model. This observation highlights a key challenge with VQ for diffusion: the VQ-compressed DiT is highly sensitive to reconstruction loss in each block. Even if each block is relatively well-reconstructed, the model easily accumulates errors, leading the final output to diverge considerably. Extending calibration to 10 epochs does little to improve performance or resolve the issue of error accumulation (\cf ~\cref{tab:vq}). This sensitivity arises from the iterative nature of diffusion models, which require $T$ forward passes to produce the final result. Hence, diffusion models demand more precise compression methods compared to LLMs or ViTs, which only require a single forward pass.

\begin{figure}[t]
    \centering
    \includegraphics[width=0.45\textwidth]{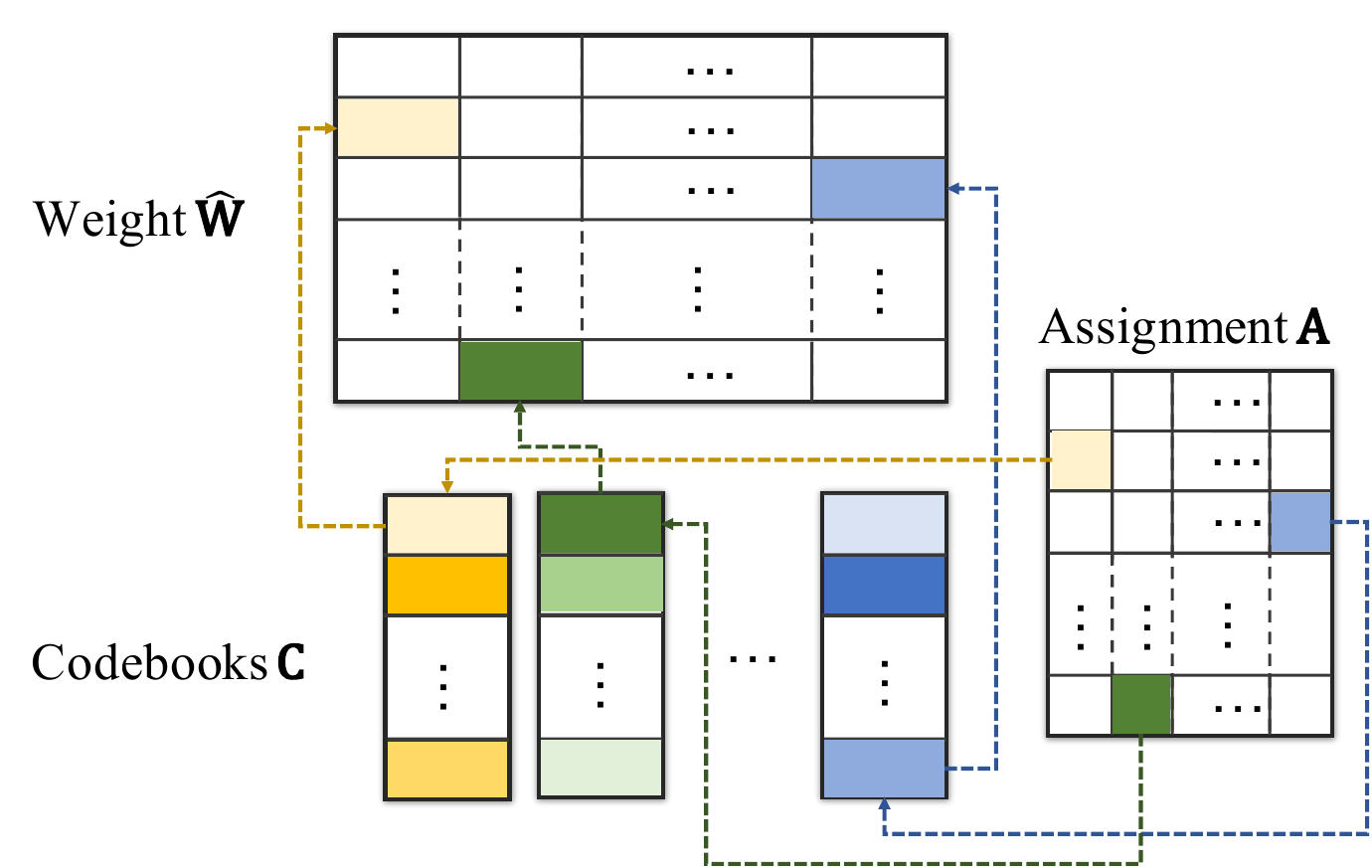}
    \caption{Product Quantization} 
    \label{codebook}
\end{figure}

\subsection{Diffusion Product Quantization}

\begin{figure*}
\centering
\includegraphics[width=0.95\textwidth]{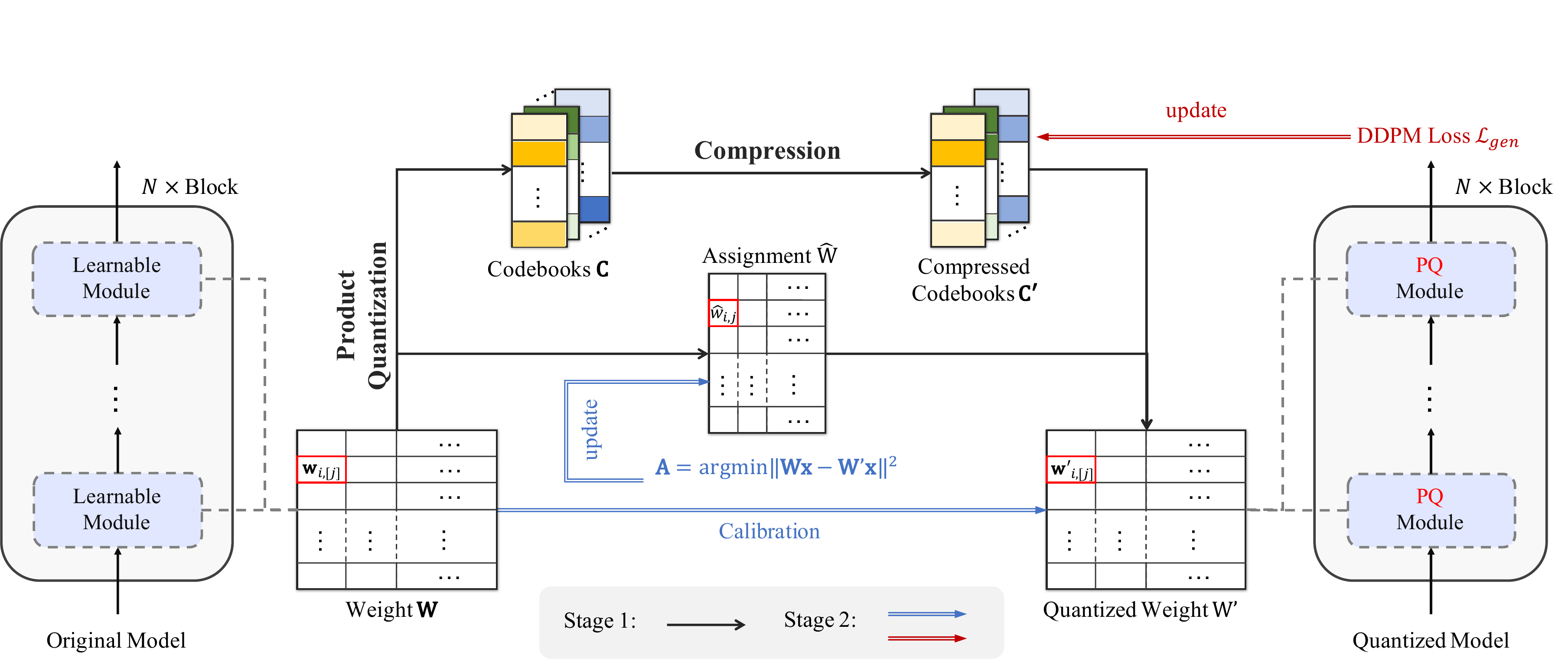}
\caption{Our method, DPQ, compresses all learnable parameters to extremely low-bits. In stage 1, we quantize each group of weights using an integer that represents a specific centroid vector in the group’s individual codebook. These codebooks are further compressed by a codebook pool, ensuring efficient model size reduction. In stage 2, we calibrate all compressed parameters by adjusting the codewords based on the reconstruction loss of the output features in the forward pass, and by fine-tuning the codebooks in the backward pass.} 
\label{fig:main}
\end{figure*}

Since diffusion models require more precise reconstruction and a larger codebook to maintain quality at lower bit rates, PQ compression is highly suitable. Product quantization (PQ) addresses the limitations of vector quantization (VQ) when an exponentially large number of codewords is needed. The core idea behind PQ is to decompose the original vector space into a Cartesian product of $N = n/d$ low-dimensional subspaces, quantizing each subspace into $k$ \emph{different} codewords. This approach yields an effective number of $k^N$ codewords in the original space, while the storage cost remains only $O(nk)$ for the codebook. The PQ method, illustrated in~\cref{codebook}, employs a codebook $\mathbf{C} \in \mathbb{R}^{N \times k \times d}$. For each sub-vector in the original weight matrix, an $\text{INT8}$ value represents its assignment, calculated as follows
\begin{equation}
    \hat{w}_{i,j} \triangleq \mathop{\arg\min}\limits_{p} \|\mathbf{w}_{i,j} - \mathbf{c}_{j,p}\|^2 \,.
\end{equation}

The storage cost of the codebook in PQ, however, is much larger than that in VQ. The size of the assignments remains the same at $m \times n / d$, but the codebook size increases from $k \times d$ to $k \times n$. Compared to the codes, the storage of the codebook in PQ is \emph{no longer negligible}, an issue that does not matter in existing PQ methods but is \emph{vital in compressing diffusion models}.

\textbf{Codebook compression.} To minimize the impact on the compression ratio, we aim for the codebook to be no more than one-quarter the size of the assignments. To achieve this, we first use FP16 to store the codebook instead of FP32, which does not affect the generation results. Building on this, we propose a new method called the ``codebook pool,'' with dimensions $N' \times d$. By calculating the memory requirements of the quantized weight $\hat{\textbf{W}}$ ($m \times \frac{n}{d} \times 8$ bytes) and the codebook $\textbf{C}'$ ($N' \times d \times 16$ bytes), we derive the target codebook size as $N' = \frac{mn}{8d^2}$. However, the projection from the old codebook to the new one also requires additional storage. After detailed calculations (see appendix), we find that the storage required for the projection is at most equal to the storage needed for the codebook pool. Therefore, we set $N' = \frac{mn}{16d^2}$ to ensure the codebook is compressed to the desired size while leaving some margin.

We observe that the original codebook $\textbf{C}$ contains many similar vectors. To address this, we define a threshold $\tau_c$ for each sub-vector: if the average L2 distance between two sub-vectors is smaller than $\tau_c$, we consider them close and eligible for merging. A simple experiment shows that directly merging close sub-vectors with the same value does not significantly affect the generation output. Previously, due to the constraints of INT8 assignments, it was difficult to merge centroids because the search space for sub-vectors in the weight was limited. However, this is no longer an issue thanks to our codebook pool strategy.

Our codebook pool is constructed as follows. We first calculate the importance of each sub-vector $\mathbf{c}_{jk}$ in the codebook, which is determined by how many codes are assigned to it. We then sort all the vectors in the codebook by importance, prioritizing the most important ones. For subsequent vectors, we check whether they are close to any vector in the codebook pool. If the pool is not yet full and the vector is not close to any existing vectors, it is added to the pool. After completing this process, if the pool is still not full, we select additional vectors from the rest, sorted by importance, to fill the pool. Our codebook compression can be described as
\begin{align}
    \mathbf{C}' &= \left\{ \mathbf{c}_{j} \;\middle|\; \underset{I(\mathbf{c}_{i}) > I(\mathbf{c}_{j})}{\min} \|\mathbf{c}_{i} - \mathbf{c}_j\|_2 > \tau_c \right\} \,, \\
    \hat{c}_{i,j} &= \mathop{\arg\min}\limits_{\mathbf{c}'_p \in \mathbf{C}'} \left\{ \left( I(\mathbf{c}_{i,j}) - I(\mathbf{c}'_{p}) \right) \;\middle|\; \|\mathbf{c}_{i,j} - \mathbf{c}'_p\|_2 < \tau_c \right\} \,,
\end{align}
where $I(\cdot)$ denotes the importance score of one centroid. After compression, we store a compressed codebook $\textbf{C}' \in \mathbb{R}^{N' \times d}$, while the original codebook is transformed into $\hat{\mathbf{C}} \in \mathbb{N}^{n/d \times k \times 1}_0$, with each integer stored in INT16 format. This codebook compression, combined with PQ, effectively expands the valid space compared to VQ, all while requiring minimal additional storage. As shown in~\cref{tab:vq}, our method adopts the codebook pool and achieves compression without significantly increasing the model size, while also extending the codebook space.

\subsection{Model Calibration}

Calibrating the compressed model is essential for preserving performance, especially for diffusion models, which are highly sensitive to computational errors. We propose  an end-to-end calibration approach similar to our implementation for vector quantization (VQ), where we calculate the DDPM loss, as shown in \cref{eq:ddpm}. This involves using the mean squared error to predict noise and back-propagating the loss to update the centroids in the codebook. An alternative approach is block-wise distillation, as used by \cite{vq4dit}, but this method has a suboptimal optimization target and cannot prevent loss accumulation. Further ablation studies will confirm this observation.

In the calibration step, the reassignment of codes is often omitted~\cite{pqf, quip}. Upon re-evaluating these VQ methods, we found that reassigning codes during calibration is largely unnecessary, as $\arg \min_{p} \|\mathbf{w}_{i,j} - \mathbf{c}_p\|^2$ remains almost unchanged. Experimental results indicate that reassigning codewords provides no substantial benefit for VQ. However, this conclusion may differ for PQ, as PQ can achieve a high level of reconstruction precision, and reassignment may further leverage the high-dimensional properties of PQ. Therefore, we argue that \emph{we should update the codewords in PQ in calibration}. Additionally, we consider the current optimization target for PQ, $\min_{p} \|\mathbf{w}_{i,j} - \mathbf{c}_{j,p}\|^2$, to be suboptimal. The primary focus of our analysis is the error in the output features, which suggests that the optimization target should be refined as follows:
\begin{equation}
    \min \|\mathbf{W}\mathbf{x} - \mathbf{W'}\mathbf{x}\|^2 \,.
    \label{eq:assign-target}
\end{equation}
where $\mathbf{W}'$ denotes the reconstructed weight. In the forward process, we gather the immediate activations and the original weights to reassign the codewords, updating the assignment by optimizing the target in~\cref{eq:assign-target}. Specifically, in each iteration, \(\hat{w}_{i,j}\) is reassigned as
\begin{equation}
    \hat{w}_{i,j} = \mathop{\arg\min}\limits_{p} \|\mathbf{w}_{i,j}\mathbf{x}_j - \mathbf{c}_{j,p}\mathbf{x}_j\|^2 \,,
\end{equation}
where \(\mathbf{x}_j\) represents the \(j\)-th group of rows in the input features. The term \(\|\mathbf{w}_{i,j}\mathbf{x}_j - \mathbf{c}_{j,p}\mathbf{x}_j\|^2\) is a \(k\)-dim vector representing the reconstruction loss of the output features. We can directly adjust the assignment based on this loss.

The calibration involves two main steps. During the forward pass, we adjust the codewords based on the reconstruction loss \(\mathcal{L}_{\text{rec}} = \|\mathbf{W}\mathbf{x} - \mathbf{W}'\mathbf{x}\|^2\). In the backward pass, we update the codebook using the DDPM loss \(\mathcal{L}_{\text{ddpm}}\) and backpropagation. The calibration step can be summarized as:
\begin{equation}
\begin{cases}
    \hat{\mathbf{W}} \leftarrow \arg \min \mathcal{L}_{\text{rec}}, & \text{forward} \\[8pt]
    \mathbf{C} \leftarrow \mathbf{C} - u \left( \frac{\partial \mathcal{L}_{\text{ddpm}}}{\partial \mathbf{C}}, \Theta \right), & \text{backward} \,.
\end{cases}
\end{equation}

The overall process is illustrated in~\cref{fig:main}. We divide the process into two main parts. The first part is data-free, where we use PQ to compress the weights and then compress the codebook to remove redundancy. The second part involves calibration using training data. In the forward pass, we dynamically calibrate the assignments based on activations, while in the backward pass, we optimize and update the codebook using the DDPM loss. 

Together, our new compression and calibration strategies revamps product quantization to enable effective Diffusion Product Quantization.

\section{Experiments}

\begin{figure*}
    \centering
    \includegraphics[width=0.9\linewidth]{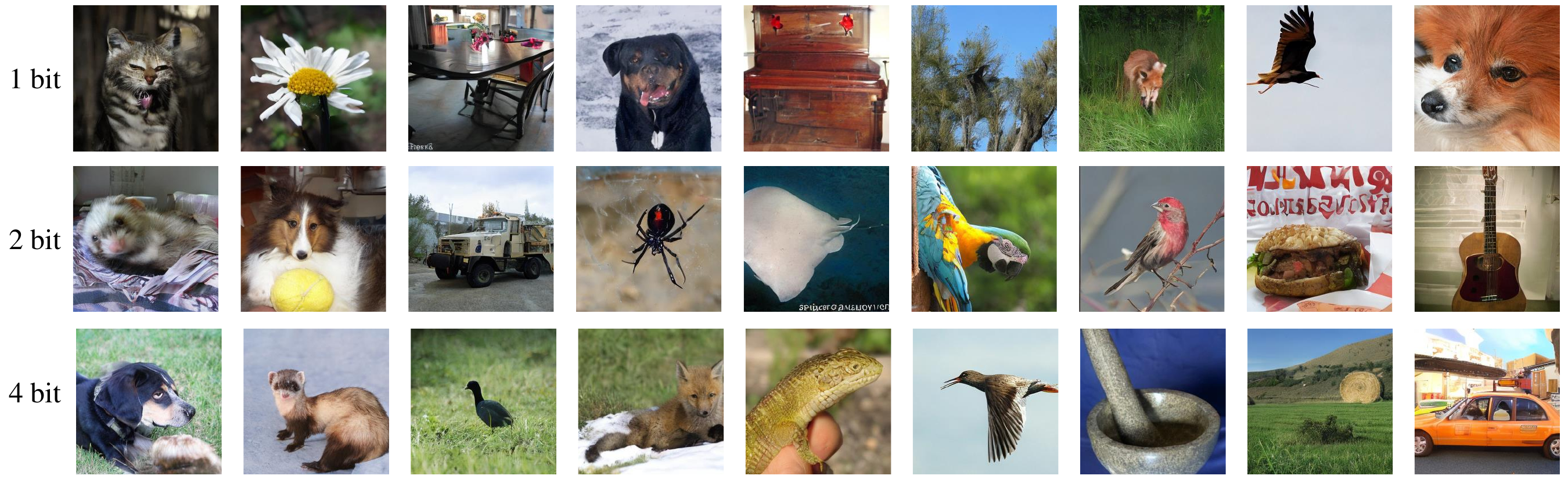}
    \caption{Visualization of generation results from the DPQ-compressed model.}
    \label{fig:visual}
\end{figure*}

\subsection{Experimental Settings}

\textbf{Models.} We conduct our experiments primarily on the advanced and widely used DiT model~\cite{dit}. The experimental setup closely follows the configuration of the original DiT setting. We use the largest pre-trained DiT-XL/2 model as our floating-point reference, serving as an upper bound. Most experiments are conducted at an image resolution of \(256 \times 256\), with results for \(512 \times 512\) resolutions provided in the appendix. To further examine the effectiveness of our method, we also perform a series of experiments on classical U-Net diffusion models.

\textbf{Metrics.} The quantized models are evaluated on the ImageNet validation set, where we typically sample 10,000 images for assessment. We use a DDPM scheduler with 50, 100, and 250 steps, along with a classifier-free guidance (CFG) of 1.5. To demonstrate the robustness of our method, we present compression results across various generation settings. For consistency with other baseline models, we limit quantization to the DiT blocks, as they contain most parameters in the diffusion model.

To evaluate the generation quality of the quantized model, we follow previous literature and employ four metrics: Fréchet Inception Distance (FID)~\cite{gan}, spatial FID (sFID)~\cite{sfid}, Inception Score (IS)~\cite{is-1,is-2}, and precision. These metrics are computed using ADM's TensorFlow evaluation toolkit~\cite{is-1}. We also report the model's compression ratio in the results section. Due to the presence of uncompressed modules and codebooks, the bit-width listed in the table does not represent the actual bits per value, especially at lower bit-widths. For a fair comparison, we present the effective compression ratio.

\textbf{Quantization.} We set \( d = 2, 3, 4, 8 \) for 4-, 3-, 2-, and 1-bit compression, respectively. $\tau_c$ is set to be 0.05 and other hyper-parameters of compression can be derived based on \( d \). The codebook and codewords are initialized using K-means with 20 iterations and our PQ implementation is built on the code released by~\cite{pqf}. For codebook fine-tuning, we use an AdamW optimizer with a constant learning rate of \( 1 \times 10^{-4} \). By default, we calibrate the DiT model for 5 epochs. Compared to the 1300 epochs required for training DiT, the calibration duration is 260 times shorter. The calibration implementation is based on the code released by~\cite{dit}. The quantization stage takes approximately 10 minutes on a single NVIDIA 3090 GPU, and the calibration stage takes around 5 hours on \( 8 \times \) NVIDIA 3090 GPUs.

\begin{table}[tb]
    \caption{Performance of our implementation of VQ and the proposed DPQ method on ImageNet at \( 256 \times 256 \) resolution, using the DiT-XL/2 model with 250 sampling steps and a CFG scale of 1.5. `Bit-width' represents the precision of quantized weights.}
    \label{vqvspq}
    \centering
    \small 
    \setlength{\tabcolsep}{3pt} 
    \resizebox{\linewidth}{!}{%
    \begin{tabular}{c|c|cccccc}
    \toprule
    Bit-width & Method & Size ratio & FID \(\downarrow\) & IS \(\uparrow\) & Precision \(\uparrow\) & Recall \(\uparrow\)\\
    \midrule[0.3pt]
    32 & FP  &  - & \pt5.33 &  275.13  & 0.8216  &  0.5790   \\
    \midrule[0.3pt]
    \multirow{2}*{4}  & VQ & \pt7.51$\times$ & \pt6.26 & 217.71 & 0.7547 & 0.6943  \\
                     & DPQ & \pt7.21$\times$ & \pt6.25 & 215.12 & 0.7576 & 0.6842  \\
    \midrule[0.3pt]
    \multirow{2}*{3}  & VQ & 10.88$\times$ & \pt7.20 & 175.23 & 0.7533 & 0.6823  \\
                     & DPQ & \pt9.92$\times$ & \pt7.01 & 188.52 & 0.7403 & 0.6868  \\
    \midrule[0.3pt]
    \multirow{2}*{2}  &  VQ & 14.87$\times$ & 12.34 & 105.67 & 0.6750 & 0.5910   \\
                     & DPQ & 12.08$\times$ & \pt6.84 & 190.99 & 0.7685 & 0.6588  \\
    \midrule[0.3pt]
    \multirow{2}*{1}  &  VQ  & 28.65$\times$ &  91.71 & \pt13.51 & 0.2481 & 0.4110    \\
                     & DPQ & 24.57$\times$ & 14.03 & 110.23 & 0.6863 & 0.6652    \\
    \bottomrule
    \end{tabular}}
\end{table}

\subsection{Experimental Results}

\begin{table*}[tb]
    \caption{Performance comparison on ImageNet at a resolution of \( 256 \times 256 \). 'Timesteps' represents the sampling steps of DiTs, and 'Bit-width' indicates the precision of quantized weights. The best result is highlighted in bold.}
    \label{timestep256}
    \centering
    \footnotesize 
    \setlength{\tabcolsep}{9pt} 
    \renewcommand{\arraystretch}{0.9} 
    \begin{tabular}{c|c|l|l|cccc}
    \toprule
    Timesteps & Bit-width & Method & Size ratio & FID \(\downarrow\) & sFID \(\downarrow\) & IS \(\uparrow\) & Precision \(\uparrow\) \\
    \midrule[0.3pt]
    \multirow{9}*{250}   & 32 & FP & - & \pt5.33 & 17.85 & 275.13 & 0.8216  \\
    \cmidrule[0.3pt]{2-8}
    & \multirow{4}*{3}   & GPTQ & 10.10 $\times$ & 50.94 & 38.60 & \pt46.37 & 0.3932 \\
                         && Q-DiT & 10.23 $\times$ & 1.3e2 & 95.86 & \pt12.75 & 0.1451 \\
                         && VQ4DiT & 10.59 $\times$ & 10.59 & 23.03 & / & / \\
                         && \textbf{Ours} & \pt9.92 $\times$ & \pt\textbf{7.01} & \textbf{19.63} & \textbf{188.52} & \textbf{0.7403}  \\
    \cmidrule[0.3pt]{2-8}
    & \multirow{4}*{2}   & GPTQ   & 14.77 $\times$ & 2.8e2 & 1.4e2 & \ptt3.97 & 0.0326  \\
                         && Q-DiT  & 15.17 $\times$ & 3.1e2 & 2.1e2 & \ptt1.28 & 0.0001  \\
                         && VQ4DiT &   15.75 $\times$ & 11.87 & 23.27 & / & /  \\
                         && \textbf{Ours} & 12.08$\times$ & \pt\textbf{6.84} & \textbf{18.75} & \textbf{190.99} & \textbf{0.7685}  \\
    \midrule[0.3pt]
    \multirow{5}*{100}   & 32 & FP  &  - & \pt5.59 & 18.63 & 269.67 & 0.8156   \\
    \cmidrule[0.3pt]{2-8}
    & \multirow{4}*{2}   & GPTQ   &   14.77 $\times$ & 2.7e2 & 1.3e2 & \ptt4.17 & 0.0001   \\
                         && Q-DiT  &  15.17 $\times$  & 3.1e2 & 2.1e2 & \ptt1.28 & 0.0003  \\
                         && VQ4DiT & 15.75 $\times$ & 11.85 & 23.64 & / & /  \\
                         && \textbf{Ours} & 12.08$\times$ &  \pt\textbf{8.18} & \textbf{20.40} & \textbf{179.69} & \textbf{0.7460}      \\
    \midrule[0.3pt]
    \multirow{5}*{50}  & 32 & FP  &  - & \pt6.72 & 21.13 & 243.90 & 0.7848   \\
    \cmidrule[0.3pt]{2-8}
    & \multirow{4}*{2}   & GPTQ &  14.77 $\times$ & 3.1e2 & 1.5e2 & \ptt4.33 & 0.0006   \\
                         && Q-DiT  & 15.17 $\times$  & 3.1e2 & 2.1e2 & \ptt1.28 & 0.0005 \\
                         && VQ4DiT &  15.75 $\times$ & 12.42 & 25.16 & / & / \\
                         && \textbf{Ours} & 12.08$\times$ & \textbf{11.04} & \textbf{23.92} & \textbf{157.61} & \textbf{0.7194} \\
    \bottomrule
    \end{tabular}
\end{table*}

\begin{table*}
    \centering
    \begin{minipage}{0.32\textwidth}
        \centering
        \captionof{table}{DDIM product quantization}
        \renewcommand{\arraystretch}{1.2} 
        \resizebox{\textwidth}{!}{
        \begin{tabular}{l|c|c|c|c}
            \toprule
            Method & Bits & Size ratio & FID & IS \\
            \midrule
            Full & 32 & - & 4.22 & 9.12 \\
            DPQ   & 4  & \pt7.76 & 5.19 & 8.62 \\
            DPQ   & 2  & 15.55 & 7.01 & 6.42 \\
            \bottomrule
        \end{tabular}
        \label{tab:ddim}
        }
    \end{minipage}
    \hfill
    \begin{minipage}{0.32\textwidth}
        \centering
        \captionof{table}{Ablation study of each module}
        \renewcommand{\arraystretch}{1.2}
        \resizebox{\textwidth}{!}{
        \begin{tabular}{c|c|c|c|c}
            \toprule
            \multirow{2}{*}{PQ} & \multirow{2}{*}{DDPM loss} & Blockwise & Codewords & \multirow{2}{*}{FID} \\
             &  & Distillation & Adjustment & \\
            \midrule
             $\checkmark$ & $\checkmark$ & & & 7.21 \\
             $\checkmark$ & & $\checkmark$ & & 8.52 \\
             $\checkmark$ & $\checkmark$ & & $\checkmark$ & 6.84 \\
            \bottomrule
        \end{tabular}
        \label{tab:ablation}
        }
    \end{minipage}    
    \hfill
    \begin{minipage}{0.32\textwidth}
        \centering
        \small
        \captionof{table}{FID scores across epochs}
        \begin{tabular}{l|c|c}
            \toprule
            Method & Epoch & FID \\
            \midrule
           \multirow{3}{*}{DPQ} & 1 & 7.32 \\
            & 5 & 6.84 \\
            & 10 & 6.79 \\
            \bottomrule
        \end{tabular}
        \label{tab:epoch}
    \end{minipage}
\end{table*}

\textbf{DPQ vs. VQ.} \Cref{vqvspq} presents a performance comparison between our implementation of VQ and the proposed DPQ method, evaluated on the DiT-XL/2 model on ImageNet at a \( 256 \times 256 \) resolution with 250 sampling steps. Across various bit-widths, DPQ consistently outperforms VQ, especially as bit-width decreases. The most significant improvement is observed at 1-bit precision: while VQ suffers from a notable collapse in performance, with an FID dropping to 91.71, DPQ achieves a much improved FID of 14.03, along with significantly higher IS (110.23 vs. 13.51) and Precision (0.6863 vs. 0.25). This demonstrates that our DPQ method effectively handles high-dimensional sub-vector compression and is suitable for extremely low-bit settings. Under a more typical 2-bit compression setting, our model also significantly surpasses VQ, improving the FID by 5.5 and increasing the IS by 85.32. As the bit-width increases, the performance gap narrows but remains still. While DPQ has a slightly larger size ratio due to the additional codebook size, this modest cost yields substantial performance gains, making it a worthwhile trade-off. Images generated by our DPQ-compressed models across different bit-widths are shown in~\cref{fig:visual}.

\textbf{Comparison with PTQ methods.} We select GPTQ~\cite{gptq}, Q-DiT~\cite{qdit}, and VQ4DiT~\cite{vq4dit} as our baselines, which are advanced post-training quantization (PTQ) techniques for DiTs. \Cref{timestep256} shows a comparison to DiT XL/2 on the ImageNet \( 256 \times 256 \) dataset. Across different timesteps (250, 100, and 50) and weight bit-widths, our method consistently achieves lower FID and sFID values compared to GPTQ, Q-DiT, and VQ4DiT, indicating improved generative quality and image fidelity. Notably, for all quantization settings, our DPQ significantly outperforms GPTQ and VQ4DiT, exhibiting substantially higher FID and sFID scores. This demonstrates DPQ's effectiveness in maintaining model performance under low bit-width quantization. Our method consistently outperforms VQ4DiT across various experimental settings, achieving FID and sFID results closer to those of the FP32 model.

We also apply our method to the traditional U-Net structure. Following~\cite{q-diffusion}, we implement product quantization on DDIM~\cite{ddim} using the CIFAR-10 dataset. \Cref{tab:ddim} illustrates that our DPQ method successfully compresses the CNN-based diffusion architecture, demonstrating its general applicability to various diffusion models.

\subsection{Ablation Study}

We evaluate the contributions of each module in our method. Since PQ alone cannot compress the model while maintaining the generation of meaningful images, we enable PQ by default. Our experiments in~\cref{tab:ablation} indicate that the DDPM loss is more effective for calibrating the codebook and achieving improved performance. Additionally, adjusting the codewords during the forward pass also enhances the performance of the compressed model.

During calibration, we observe that longer training epochs benefit our DPQ method, prompting us to investigate the optimal number of epochs. Results in~\cref{tab:epoch} indicate that FID improves significantly in the early stages, demonstrating the effectiveness of our calibration method. However, after 5 epochs, the performance gains become minimal, suggesting that the calibration effect has reached saturation at that point.

\section{Conclusion and Limitations}

In this work, we introduced Diffusion Product Quantization (DPQ), a novel framework for compressing diffusion models to extremely low bit-widths. Our analysis identifies two primary challenges when applying quantization to diffusion models: the difficulty in representing high-dimensional vectors accurately at low bit-widths and the tendency for quantization errors to accumulate over multiple diffusion steps. To address these issues, we revamped and redesigned product quantization (PQ) to allow for precise quantization with minimal codebook size. We propose a method to compress the PQ codebooks based on the usage frequency of centroids, and proposed a calibration process that adjusts assignments according to activations and updates codebooks based on the diffusion loss. Experimental results demonstrate that our DPQ method effectively compresses DiT weights to as low as 1-bit precision while preserving high-quality image generation capabilities.

There are two main limitations of DPQ. First, while DPQ reduces model size significantly, it does not directly accelerate inference. Second, the quantization process requires a considerable amount of time due to the calibration steps. We aim to address these limitations in future work.

{
    \small
    \bibliographystyle{ieeenat_fullname}
    \bibliography{main}
}


\end{document}